\documentclass[lettersize,journal]{IEEEtran}
\usepackage{amsmath,amsfonts}
\usepackage{algorithmic}
\usepackage{algorithm}
\usepackage{array}
\usepackage[caption=false,font=normalsize,labelfont=sf,textfont=sf]{subfig}
\usepackage{textcomp}
\usepackage{stfloats}
\usepackage{url}
\usepackage{verbatim}
\usepackage{graphicx}
\usepackage{cite}
\usepackage{color}
\usepackage{times}
\usepackage{soul}
\usepackage{url}
\usepackage[hidelinks]{hyperref}
\usepackage[utf8]{inputenc}
\usepackage[small]{caption}
\usepackage{graphicx}
\usepackage{amsmath}
\usepackage{amsthm}
\usepackage{booktabs}
\usepackage{algorithm}
\usepackage{algorithmic}
\begin{document}
	
	\title{STEdge: Self-training Edge Detection with Multi-layer Teaching and Regularization}
	
	\author{
		Yunfan Ye$^{+}$\thanks{+ Joint first authors. },
		Renjiao Yi$^{+}$,
		Zhiping Cai$^{*}$,
		Kai Xu$^{*}$\thanks{* Joint Corresponding authors. }
		\\
		National University of Defense Technology\\
		
			\thanks{This paper was produced by the IEEE Publication Technology Group. They are in Piscataway, NJ.}
			\thanks{Manuscript received April 19, 2021; revised August 16, 2021.}}
		
		\markboth{Journal of \LaTeX\ Class Files,~Vol.~14, No.~8, August~2021}%
		{Shell \MakeLowercase{\textit{et al.}}: A Sample Article Using IEEEtran.cls for IEEE Journals}
		
		
		\maketitle

		\begin{abstract}
			Learning-based edge detection has hereunto been strongly supervised with pixel-wise annotations which are tedious to obtain manually. We study the problem of \emph{self-training edge detection}, leveraging the untapped wealth of large-scale unlabeled image datasets. We design a self-supervised framework with multi-layer regularization and self-teaching. In particular, we impose a consistency regularization which enforces the outputs from each of the multiple layers to be consistent for the input image and its perturbed counterpart. We adopt L0-smoothing as the ``perturbation'' to encourage edge prediction lying on salient boundaries following the cluster assumption in self-supervised learning. Meanwhile, the network is trained with multi-layer supervision by pseudo labels which are initialized with Canny edges and then iteratively refined by the network as the training proceeds. The regularization and self-teaching together attain a good balance of precision and recall, leading to a significant performance boost over supervised methods, with lightweight refinement on the target dataset. 
			Through extensive experiments, our method demonstrates strong cross-dataset generality and can improve the original performance of edge detectors after self-training and fine-tuning.
			
		\end{abstract}
		
		\section{Introduction}
		Edge detection is a fundamental low-level task in computer vision, which aims to extract object boundaries and visually salient edges from natural images. Various high-level tasks have greatly benefited from edge detection, such as object detection and segmentation~\cite{li2021looking, zhu2021supplement, li2021semisupervised, arbelaez2010contour, rother2004grabcut, li2020segmenting}, video interpolation~\cite{zhao2022edge}, image inpainting~\cite{nazeri2019edgeconnect} and denoising~\cite{fang2020multilevel}.
		
		Traditional methods accentuate edges based on local features such as gradients~\cite{canny1986computational,arbelaez2010contour, dollar2014fast}. Recently, deep learning approaches have achieved great success due to their ability of capturing more global context and hence producing more meaningful edges. Examples include HED~\cite{xie2015holistically}, RCF~\cite{liu2017richer}, BDCN~\cite{he2019bi}, DexiNed~\cite{poma2020dense}, PiDiNet~\cite{su2021pixel} and EDTER~\cite{pu2022edter}.
		
		A downside of supervised learning based edge detection is the requirement of large amount of pixel-level annotations, which is extremely tedious to obtain manually. Meanwhile, manual labels are biased from person to person, the Multicue~\cite{mely2016systematic} dataset were labeled by six annotators whose annotations are often inconsistent and the ground-truths were obtained by taking the average. 
		To this end, we study the problem \emph{self-supervised edge detection}, leveraging the untapped wealth of large-scale unlabeled image dataset. In fact, we found that self-training is especially suited for edge detection when reasonably designed and trained. Highly accurate and generalizable models can be learned with a large collection of unlabeled data.
		
		We design a network that is self-trained with multi-layer regularization and self-teaching. This multi-layer self-training is inspired by the supervised HED method~\cite{xie2015holistically} which shows that edge detection networks are best learned with deep supervision on multiple layers. For the purpose of self-training, we impose \emph{consistency regularization}~\cite{french2019semi, kim2020structured} which enforces the output from each of the multiple layers to be consistent for the input image and its perturbed counterpart. In particular, we adopt L0-smoothing~\cite{xu2011image} as the ``perturbation'' to encourage edge prediction lies in salient boundaries. This conforms with the cluster assumption in generic self-supervised learning~\cite{chapelle2005semi} as L0-smoothing~\cite{xu2011image} is inherently a color-based pixel clustering. This way, our network learns to discriminate salient edges.
		
		Meanwhile, our network is self-trained with multi-layer teaching (supervision) by pseudo labels. The pseudo labels are initialized with Canny edges and iteratively refined by the network as training proceeds. In each iteration, the edge map output by the network is firstly binarized and pixel-wise multiplied with the low-threshold (over-detected) Canny edge map before being used as pseudo labels for the next round. This is essentially an entropy minimization which helps improve pixel classification with low density separation~\cite{grandvalet2005semi}.
		

		The multi-layer regularization and self-teaching together achieve a good balance of precision and recall of edge detection, making our method realizes good cross-dataset generalization. After self-trained on \textit{COCO} validation dataset without using labels, it attains $1.2\%$ improvement for ODS and $1.8\%$ for OIS when tested on the unseen \textit{BIPED} dataset, compared to supervised methods trained on \textit{BSDS} dataset using all labels. 
		Also, on \textit{BIPED} dataset, based on our self-training method, two backbones of DexiNed and PiDiNet finetuned on $50\%$ of the training set already outperforms all the state-of-the-art methods finetuned on $100\%$ of the training set. With the same training set, our method improves the original backbones in all cases.


		
		In a nutshell, our contributions are mainly as follows:
		\begin{itemize}
			\item We propose the first framework which enables self-training edge detection from single images.
			\item We introduce multi-layer consistency regularization and self-teaching in the context of self-trained edge detection.
			\item We conduct extensive evaluations of the self-trained network and demonstrate its strong cross-dataset generality and the potential to promote edge detectors.
		\end{itemize}

		\section{Related Work}
		\paragraph{Edge Detection}
		Existing edge detection methods can be categorized into three groups, traditional edge detector, learning based methods and deep learning based ones.
		
		Traditional edge detectors focus on utilizing image gradients to generate edges such as Sobel~\cite{kittler1983accuracy} and Canny~\cite{canny1986computational}. Although they suffer from noisy pixels and do not consider semantic understanding, they are still widely used in applications such as image segmentation~\cite{arbelaez2010contour, rother2004grabcut} and image inpainting~\cite{nazeri2019edgeconnect}.
		
		Learning based methods usually integrate various low-level features and train detectors to generate object-level contours, based on priors such as gradient descent~\cite{arbelaez2010contour} and decision tree~\cite{dollar2014fast}. Although these methods achieve better performance than traditional edge detectors, they have many limitations in challenging scenarios. 
		
		Amounts of deep learning based methods have been proposed with the success of Convolutional Neural Network (CNN). In early stage, there are patch-based approaches like DeepEdge~\cite{bertasius2015deepedge} and DeepContour~\cite{shen2015deepcontour} which take pre-divided patches as input of CNNs to decide edge pixels. HED~\cite{xie2015holistically} is a pioneering work of end-to-end edge detection, with a network architecture based on VGG16~\cite{simonyan2014very} and parameters adopted from pre-trained models on ImageNet dataset~\cite{deng2009imagenet}. Based on HED, RCF~\cite{liu2017richer} combines richer features from each CNN layer and BDCN \cite{he2019bi} proposed a bi-directional cascade structure to train the network with layer-specific supervisions. Efforts have also been made to design lightweight architectures for efficient edge detection including \cite{wibisono2020fined, poma2020dense, su2021pixel}, where DexiNed~\cite{poma2020dense} introduces Xception~\cite{chollet2017xception} to edge detection network and PiDiNet integrates the traditional edge detection operators into CNN models. In this paper, unlike previous works, we neither focus on the network architecture design and efficiency improvements. We first introduce self-training into edge detection and explore a framework to utilize unlabeled dateset.
		
		
		\paragraph{Self-training}
		Self-training is a semi-supervised or unsupervised learning strategy that iteratively train the network with constantly updated pseudo labels for unlabeled training data. It is widely studied on image classification \cite{scudder1965probability, fralick1967learning, lee2013pseudo}, and recently applied to high-level vision tasks such as semi-supervised segmentation \cite{chen2020naive, zoph2020rethinking, zhu2020improving, hung2018adversarial, ibrahim2020semi}. However, in their cases, the pretrained models used in self-training are usually trained supervisedly on labeled data. 
		Recently, some zero-shot methods based on transferable and adversarial networks ~\cite{chang2021comprehensive, zhang2022tn, li2022video, yan2021zeronas} also show the potential for pre-training models.
		
		The most similar framework ULE~\cite{li2016unsupervised} for self-training edge detection is based on optical flow estimation, by iteratively detecting motion edges as pseudo labels from videos and re-training the edge detector. However, learning from consecutive frames need extra cost of storage and calculation, and the focus of only motion edges limits the performance. In this work, we aim to enable self-training edge detection from single images.

		To adapt with self-training on noisy pseudo labels, various kinds of perturbations and consistency methods are studied. Image perturbation methods~\cite{french2019semi, kim2020structured} augment the input images randomly and constraint the predictions of augmented images to be consistent with the original one. Feature perturbation method~\cite{ouali2020semi} uses multiple decoders and enforces the consistency between decoder outputs. Network perturbation method~\cite{ke2019dual} applies two networks of the same structure with different initialization and impose the consistency between the predictions of perturbed networks. Moreover, Chen et al. \cite{chen2021semi} imposes the consistency by using predictions of one network to supervise the other one. In this work, with the inspiration of previous methods, we also impose consistency regularization in our self-training framework. Specifically, L0-smoothing~\cite{xu2011image} is adopted to help the network learn to discriminate salient edges based on the cluster assumption, and a post-process is delicately designed for entropy minimization which helps improve pixel classification with low density separation~\cite{grandvalet2005semi}.
		
		\section{Method}
		In Section~\ref{sec:formulation}, we provide problem formulations and overview of the pipeline including a self-training scheme consisting of multi-layer teaching and consistencies. 
		In Section~\ref{sec:teaching}, we introduce multi-layer teaching by noisy pseudo labels, to enable training on unlabeled images.
		In Section~\ref{sec:regularization}, we introduce a multi-layer consistency constraint to regularize the multi-layer teaching, where we enforce the outputs from each of the multiple layers to be consistent for the input image and its smoothed counterpart. 
		In Section~\ref{sec:self training}, we introduce the iterative self-training process including post-process for entropy minimization.
		In Section~\ref{sec:retraining}, we adopt an uncertainty-aware strategy to filter pseudo labels for further retraining to avoid overfitting noisy edge pixels and thus improve the performances.
		
		\begin{figure*}
			\centering
			\includegraphics[width=\linewidth]{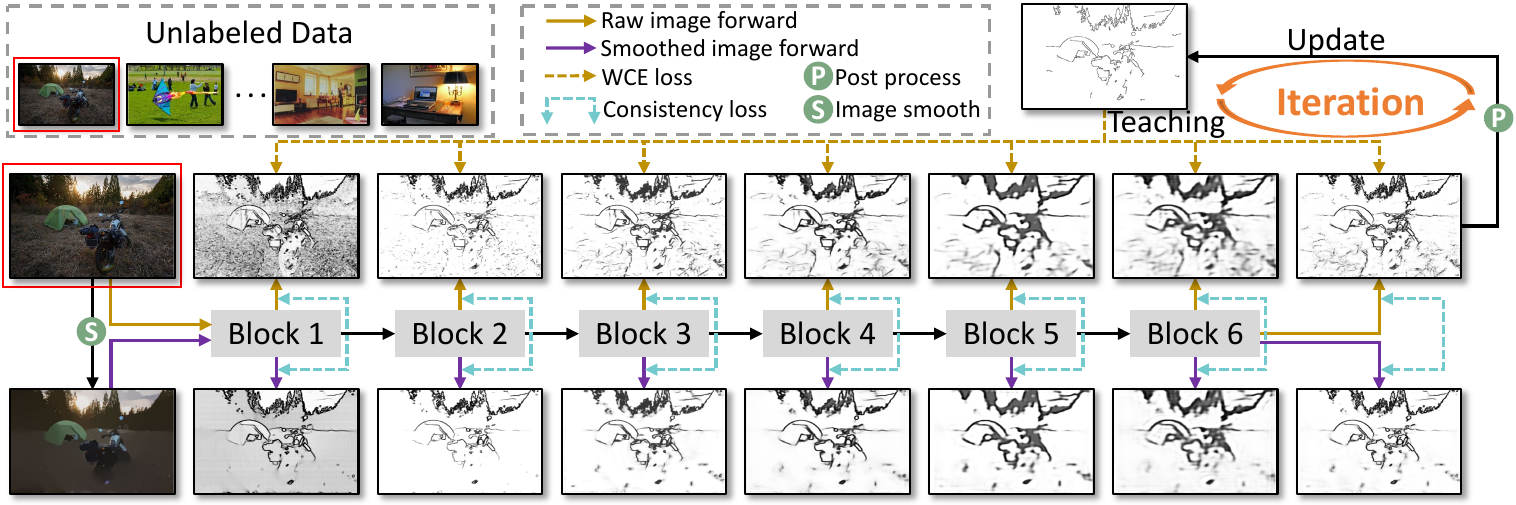}
			\caption{Our self-training framework. The input is an unlabeled image and its smoothed counterpart. Consistency regularization and teaching happen in multiple layers of their output edge maps. Pseudo labels are iteratively generated and send into the network again for self-training.}
			\label{fig:structure}
		\end{figure*}
		
		\subsection{Method Overview}\label{sec:formulation}
		We aim to propose a self-supervised training scheme to utilize unlabeled images for edge detection. 
		
		To initialize the network as our phase-one model, we use pseudo labels generated by Canny~\cite{canny1986computational} as supervision.
		We observe that even some edge pixels are missing in pseudo labels, the phase-one model can still predict themselves correctly, rather than overfitting to noisy pseudo labels, as most pixels are labeled correctly. 
		It motivates the idea of using the phase-one model as initialization, and gradually improves pseudo labels in the self-training phase.
		Actually, there are several merits of adopting pseudo label learning for edge detection. First, the pixel-level annotations for edge detection are expensive and tedious to get by brute force. As a result, there are few edge detection datasets available. Also, scale of current datasets with annotations are quite small, leading overfitting in supervised training. Naturally, it motivates our work on utilizing large-scale unlabeled image datasets by self-training scheme. Moreover, despite that pseudo labels inevitably contain noises, most of them are correct enough. By introducing multi-layer consistency regularization, pseudo labels can be gradually stable and less noisy to serve as extra training data to promote the performance. 
		
		As shown in Figure~\ref{fig:structure}, 
		in the self-training phase, the pseudo labels and network predictions are improved jointly and iteratively. The pseudo labels teach the network at multiple layers with the proposed multi-layer consistency as regularization, to suppress the complicated textures and prevent the network from generating redundant edges. During the self-training, the input image $X$ and its perturbed one $X^{\prime}$, are sent into the same network $f\left(x\right)$. Multi-layer consistency losses are defined between predictions of $X$ and $X^{\prime}$. In this paper, we apply Gaussian blur followed by L0-smoothing~\cite{xu2011image} to $X$ to get $X^{\prime}$.
		The process can be logically demonstrated as below:
		\begin{equation}
			\begin{aligned}
				&X \rightarrow f\left(x\right)  \rightarrow  \hat{P} \rightarrow Y \\
				&\downarrow  \ \nearrow \qquad \ \ \searrow  \ \updownarrow \\
				&X^{\prime } \qquad\quad \ \ \ \ \ \hat{P^{\prime}}
			\end{aligned}
		\end{equation}
		$\hat{P}$ is a set of predicted edge maps generated from multiple layers of the network, $\hat{P}=[\hat{p_{1}}, \hat{p_{2}},...,\hat{p_{n}}]$, where $\hat{p}_{i}$ has the same size as input image, and $n$ is the number of outputs from each upsampling block. $\hat{P}$ and $\hat{P^{\prime}}$ are predicted edge maps of multiple layers from $X$ and $X^{\prime}$ respectively. $Y$ denotes the pseudo labels post-processed from $\hat{P}$ for the next round training based on entropy minimization.
		
		

		\subsection{Multi-layer Teaching}\label{sec:teaching}
		Deeply-supervised schemes are studied and proven to be effective on edge detection~\cite{xie2015holistically}. The receptive field varies from different network layers. Low-level layers with small receptive fields are  likely to detect fine details, while high-level layers tend to decide semantic boundaries.
		Providing teaching of edge maps at multiple layers is essential and effective (See Section~\ref{sec: ablation} for more details). Previously, edge detection networks perform better with deep supervisions at multiple layers \cite{xie2015holistically}, followed by several recent methods~\cite{liu2017richer,he2019bi,poma2020dense}. Since the distribution of edge/non-edge pixels is heavily biased in a natural image, we adopt the robust weighted cross entropy loss denoted as $\mathcal{L}_{wce}$~\cite{liu2017richer}  at multiple layers to drive the training. Loss at pixel $x_{i}$ of an up-sampling block $n$ is calculated by:
		
		\begin{equation}\label{eq:wce_loss_pixel}
			l_{wce}^{n}\left(x_{i} , W\right)=
			\left\{
			\begin{array}{lll}
				\alpha \cdot \log \left(1-P\left(x_{i} , W\right)\right), &  if \ y_{i}=0, \\
				0, & if \ 0 < y_{i} < \eta, \\
				\beta \cdot \log P\left(x_{i} , W\right), & otherwise.
			\end{array}
			\right.
		\end{equation}
		in which
		\begin{equation}
			\begin{array}{l}
				\alpha=\lambda \cdot \frac{\left|Y^{+}\right|}{\left|Y^{+}\right|+\left|Y^{-}\right|}, \\
				\beta=\frac{\left|Y^{-}\right|}{|Y+|+|Y-|},
			\end{array}
		\end{equation}
		where $W$ denotes the collection of all network parameters, $Y^+$ and $Y^-$ denote the number of edge pixels and none-edge pixels in the ground truth,  respectively. $\lambda$ is a hyper-parameter to balance positive and negative samples. $\eta$ is a threshold to filter out the less confident edge pixels in ground truths, to avoid confusing the network. Thus, loss after an upsampling block $n$ for input image $X$ of size $w \times h$ can be represented as:
		\begin{equation}
			l^{n}_{wce}\left(X, W\right)=\sum_{i=1}^{w \times h} l^{n}_{wce} \left(x_{i},W\right).
		\end{equation}
		We define different weight $\delta^{n}$ for each level, and the final $\mathcal{L}_{wce}$ is calculated as:
		\begin{equation}\label{eq:wce_loss}
			\mathcal{L}_{wce}=\sum_{n=1}^{N} \delta^{n} \times l^{n}_{wce}\left(X, W\right).
		\end{equation}

		\subsection{Multi-layer Regularization}\label{sec:regularization}
\begin{figure}
			\centering
			\includegraphics[width=1\columnwidth]{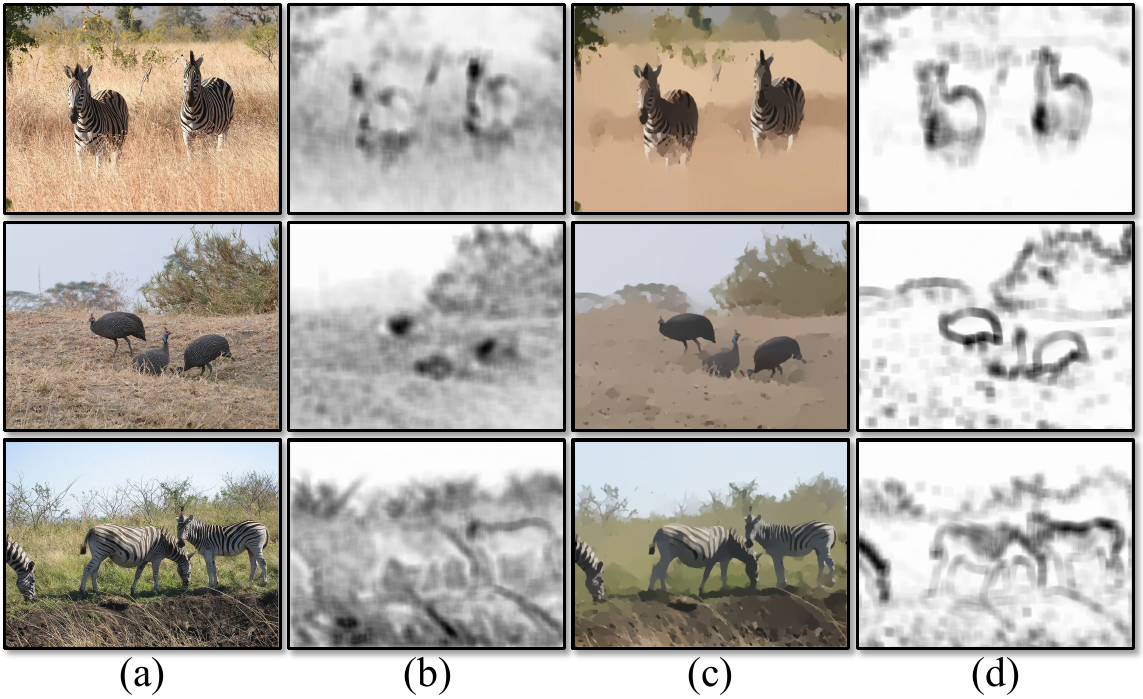}
			\caption{\textbf{The cluster assumption in edge detection.} (a) Examples from COCO dataset. (b) \emph{For raw images}. The average Euclidean distance between each patch of size 20×20 centered at a given spatial location extracted from the input images, and its 8 neighboring patches. (c) Results from applying Gaussian blur followed by L0-smoothing to raw images. (d) \emph{For smoothed images}. The average Euclidean distance map. Darker regions indicate higher distances.}
			\label{fig:cluster_assumption}
		\end{figure}

		Consistency regularization is widely studied to make the decision boundary lies in low-density areas. We enforce the consistency at multiple layers between the input image and its smoothed counterpart. Here we apply Gaussian blur followed by L0-smoothing \cite{xu2011image} to the input image, imposing consistency to help the network converge to predict more visually-salient edges. 
		As illustrated in Figure~\ref{fig:cluster_assumption}, edge pixels are easier to be classified in smoothed images which suppress complicated texture pixels. 
		
		The multi-layer consistency works as a regularizer to prevent the network from generating increasingly redundant and noisy edges as in pseudo labels. 
		Although image noises and redundant edges are expected to be filtered out after L0-smoothing as an edge-preserving smoothing approach, it unavoidably causes over-sharpening in challenging circumstances~\cite{xu2011image}. Therefore, edge maps predicted from smoothed images still contain noises and it has been proven in \cite{ghosh2017robust} that though categorical cross entropy loss converges rapidly, it is sensitive to noises. To avoid overfitting to incorrect labels during self-training, we apply L2 norm, a symmetrical loss fuction, which is more noise-tolerant on classification problems based on the theory of risk minimization~\cite{ghosh2015making,ghosh2017robust}, where edge detection is exactly a binary classification problem at pixel level. As L2 loss is theoretically and practically effective in our self-training method for its noise tolerance. 
		The pixel-wise squared difference is computed between the edge map of each block predicted from $X$ and the corresponding edge map from $X^{\prime}$. The multi-layer consistency loss for block $n$ is formulated as:
		\begin{equation}
			l^{n}_{mlc}\left(X, W\right)=\sum_{i=1}^{w \times h} \left [ P\left(x_{i} , W\right)-P\left(x_{i}^{\prime} , W\right) \right ]^{2},
		\end{equation}
		same with the $\mathcal{L}_{wce}$, the complete $\mathcal{L}_{mlc}$ with layer weights $\delta$ is calculated as:
		\begin{equation}\label{eq:mlc_loss}
			\mathcal{L}_{mlc}=\sum_{n=1}^{N} \delta^{n} \times l^{n}_{mlc}\left(X, W\right).
		\end{equation}
		Combining multi-layer teaching and regularization, with a trade-off weight $\mu$, the final loss is calculated as:
		\begin{equation} \label{eqn: whole_loss}
			\mathcal{L}= \mathcal{L}_{wce} + \mu \mathcal{L}_{mlc}.
		\end{equation}

		\begin{figure*}
			\centering
			\includegraphics[width=\linewidth]{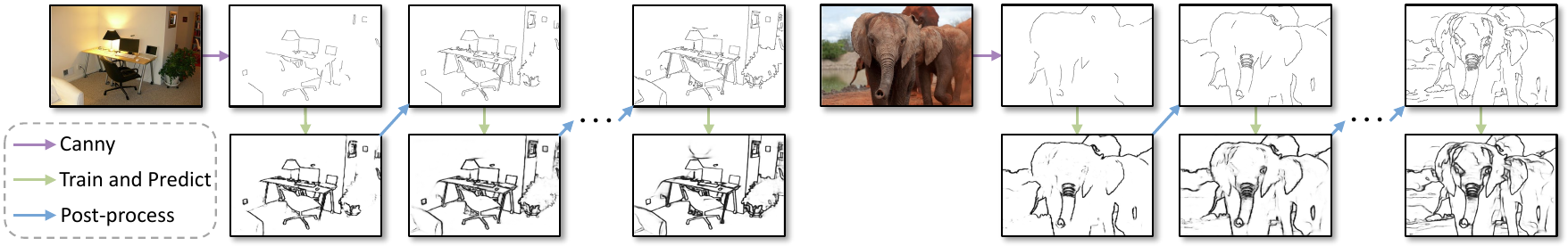}
			\caption{The evolving process of two examples during the iterative training. For each example, the top left is the input image, the first row illustrates the updating pseudo labels, and the second row shows the network predictions as iteration proceeds. }
			\label{fig:coco_evolve}
		\end{figure*}
		
		\begin{algorithm}[htb]
			\renewcommand{\algorithmicrequire}{\textbf{Input:}}
			\renewcommand{\algorithmicensure}{\textbf{Output:}}
			
			\caption{Self-training Edge Detection (\textbf{STEdge}).}
			\label{alg:Framwork}
			\begin{algorithmic}[1] 
				\REQUIRE ~~\\ 
				The unlabeled images $X$; The number of epochs trained in each round $E$; The termination parameter $T\%$.
				\ENSURE ~~\\ 
				The well-trained edge detector $M$ with parameters $\theta$;\\
				The images $X$ with reasonable edge pseudo labels $Y$. 
				
				\STATE Initialize the network weights supervisedly trained by Eq.~\ref{eq:wce_loss}. Initial pseudo labels are generated by performing Canny of high threshold on blurred images:
				
				$Y = Canny (Blur (X), thres\_high)$,
				
				$\theta_0 = Train(M(\theta), X, Y, \mathcal{L}_{wce})$;
				
				\STATE Prepare smoothed images for consistency regularization:
				
				$X^{\prime}= L0\_smoothing (Blur (X))$;
				
				\FOR{round $k$ in $\left \{ 1,...,K \right \}$ }
				\STATE Get the edge maps predicted by current weights:
				
				$\hat{P}=[\hat{p_{1}}, \hat{p_{2}},...,\hat{p_{n}}] = M(\theta_{k-1}, X)$;

				\STATE Post-process to get new pseudo labels by performing element-wise multiplication
				on over-detected Canny and our binarized edge map:  
				
				$A_k = Adaptive\_binarization(\hat{p_n})$,
				
				$C_k = Canny (Blur(X), thres\_low)$,
				
				$Y_k = Connectivity\_filter(A_k \odot C_k) $;
				
				\STATE Train on current pseudo labels for $E$ epochs 
				by Eq.~\ref{eq:wce_loss} and Eq.~\ref{eq:mlc_loss}:
				
				$\theta_k = Train(M(\theta_{k-1}), X, X^{\prime}, Y,\mathcal{L}_{wce}, \mathcal{L}_{mlc})$;
				
				\STATE Decide if the self-training process is stable and needed to be terminated, the number of edge pixels in the pseudo labels produced in round $k$ is termed $N_{edge}^{k}$: 
				
				
				\textbf{if} $\frac{N_{edge}^{i} - N_{edge}^{i-1}}{N_{edge}^{i}}< T\%$ \textbf{then}
				
				\ \ \ \ \textbf{break};
				
				\textbf{end if}. 
				\ENDFOR. 
				
			\end{algorithmic}
		\end{algorithm}
		
		\subsection{Iterative Self-training}\label{sec:self training}
		
		Our approach includes two phases, the initialization phase and the self-training phase. Phase-one training aims to get an initial model to warm up self-training, which is needless to be very powerful. In phase-one, we apply Canny of high thresholds to the unlabeled dataset $D^{u}$ to generate pseudo labels, then train the network using weighted cross entropy loss.
		
		Phase two is the iterative self-training, optimizing the network and pseudo labels simultaneously as training proceeds. Based on entropy minimization, the predicted edge maps are post-processed to be the updated labels for the next round. 
		In Algorithm~\ref{alg:Framwork}, $\hat{p_{n}}$ is the network prediction of the last layer, with values between 0 and 1 at each pixel.
		In post-processing, $\hat{p_{n}}$ is multiplied with over-detected Canny edges (with low thresholds) in a  pixel-wise level, and then filtered by connected areas with low connectivity to clear noise edges. 
		The updating of pseudo labels is critical from two aspects. On one hand, if we treat the predicted edge maps as new pseudo labels, the loss and gradients will be all-zeros. On the other hand,
		when the predicted probability map is transferred into binary edge map, the entropy of each pixel is minimized, enforcing low density seperations~\cite{grandvalet2005semi} by predicted edges.
		
		During training, we repeat the following process:
		
		(a) Predict then post-process $\hat{p_{n}}$ to generate new pseudo labels $Y$;
		
		(b) Training the network for $E$ epochs using updated pseudo labels.
		
		We refer to an iteration of (a) and (b) as one \emph{round}. The network is self-trained iteratively for several rounds until convergence. There are several factors enabling the pseudo labels to become gradually stable during self-training.
		First, in our post-processing, we apply Canny with preset thresholds to generate updated pseudo labels, which means the edge pixels have an upper bound (Canny edges on original input). Also, though Canny edges contain a lot of noisy pixels other than salient edges, our multi-layer regularization is introduced to encourage higher importance on salient edges and suppress noisy ones. With the premise of limited upper bound, as the consistency loss converges, the pseudo labels become gradually better and stable during the self-training phase. 
		
		The self-training process is terminated when the pseudo edge maps changes slightly. Specifically, we finish the self-training when the ratio of increased edge pixel number to the total edge pixel number is smaller than a termination value $T\%$ (We set $T=2$ in all experiments). We summarize the whole self-training process as in Algorithm~\ref{alg:Framwork}. The qualitative evolving process is shown in Figure~\ref{fig:coco_evolve}.

		\subsection{Uncertainty-aware Re-training}\label{sec:retraining}
\begin{figure}
			\centering
			\includegraphics[width=\linewidth]{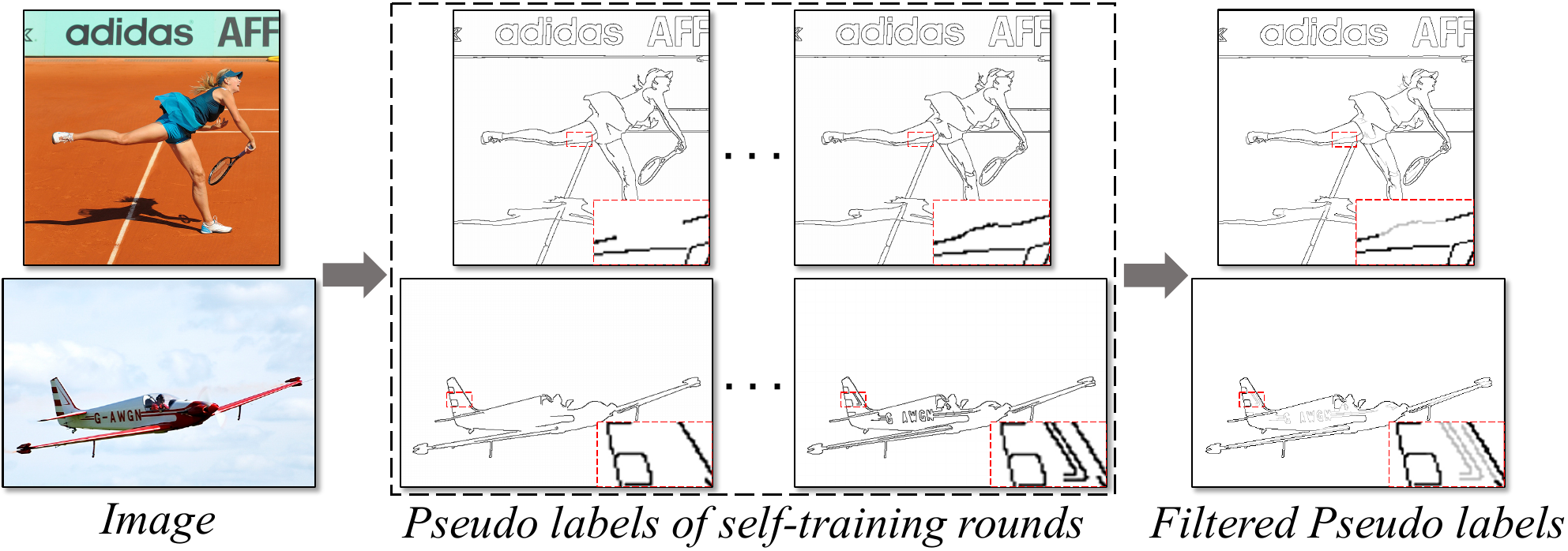}
			\caption{Two examples from \textit{COCO val2017} dataset of the uncertainty-aware pseudo label filtering process. Zoom-in is recommended for better observation. Edge pixels of low responses are uncertain pixels to be ignored when retraining with filtered pseudo labels.}
			\label{fig:filter_pseudo_labels}
		\end{figure}

		To make full use of the pseudo labels from each self-training round, and avoid overfitting noisy ones, we adopt an uncertainty-aware strategy to filter pseudo labels for further retraining. Specifically, only the edge pixels that appears in pseudo labels of every self-training round that can be treated as true edge pixels (set to 1), while other edge pixels are uncertain ones (set to a value which is smaller than $\eta$ in Equation~\ref{eq:wce_loss_pixel}) that will be ignored during the loss calculation when retraining. Examples are presented in Figure~\ref{fig:filter_pseudo_labels}.
		The filtered pseudo labels are also generated without any human annotations, which can better play as extra and free datasets to be trained by any learning-based edge detectors from the scratch.
		
		\section{Experiments}

		\subsection{Datasets}
		Datasets adopted in our experiments include \textit{COCO}~\cite{lin2014microsoft}, \textit{BSDS} \cite{arbelaez2010contour}, \textit{Multicue} \cite{mely2016systematic} and \textit{BIPED} \cite{poma2020dense}. \textit{COCO} is a widely used computer vision dataset containing common objects for natural scenes, in this paper, we adopt \textit{COCO val2017} of 5000 images as unlabeled dataset for the self-training. \textit{BSDS} is designed for image segmentation and boundary detection, consisting of 200, 100 and 200 images for training, validation and testing respectively. Each image is labeled by multiple annotators and the final ground truth is calculated by taking their average. As previous works, we augment the training and validation sets with flipping (2×), scaling (3×) and rotation (16×) during the training. It also plays as the training set of several state-of-the-art networks to compare the generalization ability. \textit{Multicue} is a commonly-used benchmark dataset for edge detection, containing 100 images from challenging natural scenes. Each scene contains a sequence of images from the left and right views, captured by a stereo camera. Only the last frames of every left-view sequences are labeled by annotators. For these 100 images, each of them are annotated by several people as well. As previous works, we randomly split them into training and evaluation sets, consisting of 80 and 20 images respectively. \textit{BIPED} dataset contains 250 annotated images of outdoor scenes, spliting into a training set of 200 images and a testing set of 50 images. All images are carefully annotated at single-pixel width. The resolution of \textit{Multicue} and \textit{BIPED} are both 1280 $\times$ 720. Thus we augment each image by flipping (2×), cropping (3×), and rotation (16×), leading to a training set that is 96 times larger than the original dataset. We conduct ablation study and the evaluation of cross-dataset generality on \textit{BIPED} dataset. The performances after finetuning on \textit{BSDS}, \textit{Multicue} and \textit{BIPED} are also compared with state-of-the-art edge detectors.

		\subsection{Implementation Details and Evaluation Metrics}
		The proposed self-training framework is implemented in PyTorch \cite{paszke2019pytorch}, and can be applied to any edge detection network with light refinements. The filtered pseudo labels can also be trained with any edge detectors from the scratch. Following previous works, the parameter $\lambda$ to balance positive and negative samples and the threshold $\eta$ in $\mathcal{L}_{wce}$ is set to 1.1 and 0.3 in all cases, respectively. In the final loss function $\mathcal{L}= \mathcal{L}_{wce} + \mu \mathcal{L}_{mlc}$, the selection of the trade-off weight $\mu$ do not bring much difference and we just set it to 1, and the termination value $T\%$ in Algorithm~\ref{alg:Framwork} is set to $2\%$ (generally 6-8 iterations). No data augmentation strategies are adopted when self-training with pseudo labels. All Experiments are conducted on an NVIDIA GeForce RTX 3080 Ti GPU with 12GB memory. 
		
		In this paper, two current lightweight edge detectors DexiNed~\cite{poma2020dense} and PiDiNet~\cite{su2021pixel} are adopted to show the effectiveness of our STEdge. Both networks are trained with the Adam optimizer~\cite{kingma2014adam}. For the backbone of DexiNed, the learning rate is 0.0001 with the batch size of 8. 
		For the backbone of PiDiNet, we train the network with the batch size of 24, decaying in a multi-step way at an initial learning rate of 0.005. 
		
		For evaluations, like previous works, standard non-maximum suppression (NMS) will be applied to thin detected edges before evaluation. We adopt three commonly-used evaluation metrics for edge detection, the F-measure of optimal dataset scale (ODS), the F-measure of optimal image scale (OIS) and average precision (AP). ODS and OIS are two strategies to transform the output probability map into a binary edge map. ODS employs a fixed threshold for all images in the dataset while OIS chooses an optimal threshold for each image.
		The F-measure is defined as $F =  \frac{2\cdot P\cdot R}{P+R}$, where $P$ denotes precision and $R$ denotes recall.
		AP represents the area under the Precision-Recall curve. However, in some cases when the Precision-Recall curve is shorter and does not cover the whole range, it gives a lower AP even if the predictions are better and satisfactory. In such cases, AP may be less reliable for evaluations compared with ODS and OIS.
		For ODS and OIS, the maximum allowed distances between corresponding pixels from predicted edges and ground truths are set to 0.0075 for all experiments.

		\subsection{Ablation Study} \label{sec: ablation}
		We evaluate the effectiveness of each part of our STEdge pipeline by conducting several ablations. 
		
		\begin{table}[htbp]
			\centering
			\scalebox{1}{
				\begin{tabular}{c|ccc}
					\hline
					Method & ODS   & OIS & AP\\
					\hline
					Canny (20, 40) & .266  & \textbackslash{} & \textbackslash{}\\
					Canny (100, 200) & .680  & \textbackslash{} & \textbackslash{}\\
					Canny (200, 300) & .597 & \textbackslash{} & \textbackslash{}\\
					Phase-one Model& .705  & .724 &  .684\\
					
					Consistency (only last-layer teaching) & .739 & .757& .697\\ 
					Multi-layer teaching & .754 & .769 & .738\\
					Multi-layer teaching + Consistency & .760 & .783 & \textbf{.798}\\
					Multi-layer teaching + Consistency + Re-training& \textbf{.788} & \textbf{.803} & .754\\
					\hline
			\end{tabular}}
			\caption{Ablation studies of several critical modules on \textit{BIPED} dataset. }
			\label{tab:ablations}
		\end{table}
		
\begin{figure}
			\centering
			\includegraphics[width=1\linewidth]{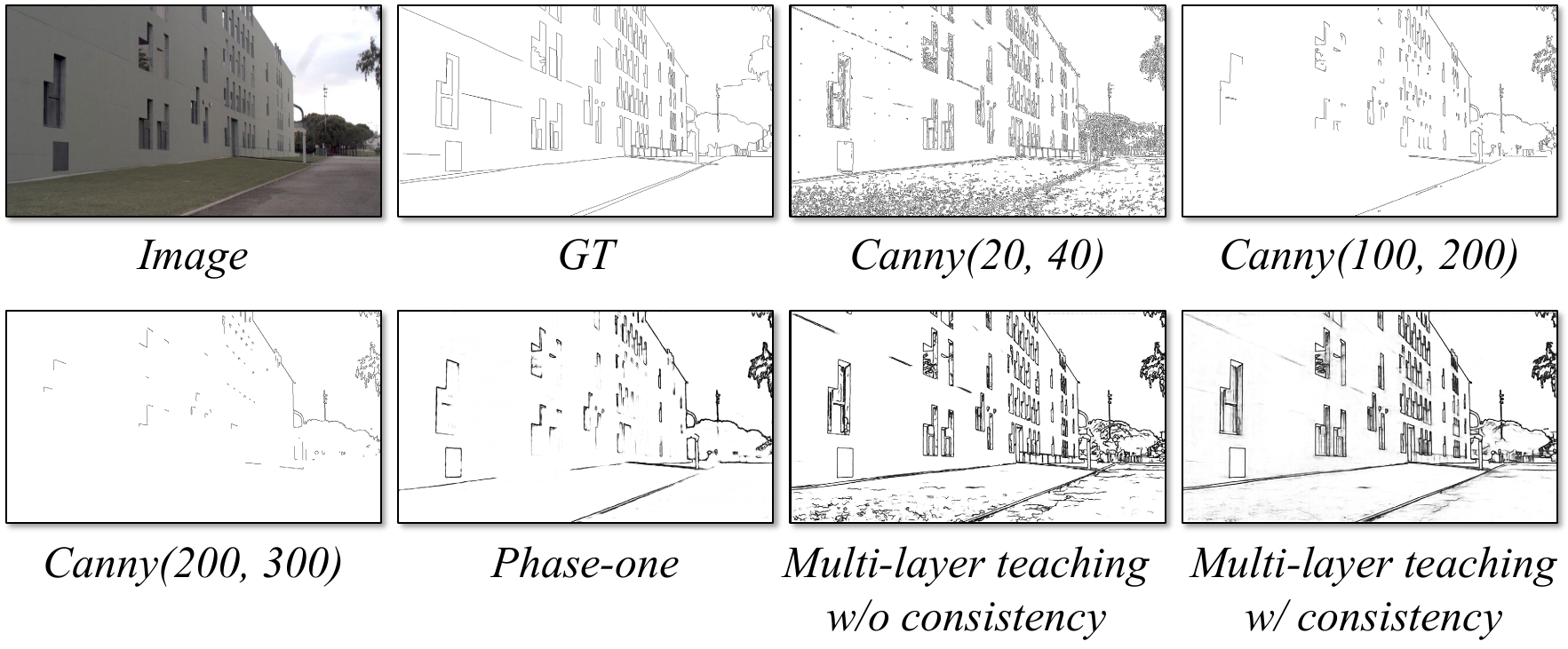}
			\caption{Edge maps from Canny detectors with different thresholds and three different settings of our STEdge on the unseen \textit{BIPED} dataset. STEdge are trained without any human annotations. }
			\label{fig:ablation_biped}
		\end{figure}

		In terms of the DexiNed backbone, Table~\ref{tab:ablations} presents the quantitative comparisons of Canny detectors with different thresholds, the phase-one model, only last-layer teaching, multi-layer teaching with and without consistency regularization and the final model retrained with filtered uncertainty-aware pseudo labels. 
		Qualitative comparisons are in Figure~\ref{fig:ablation_biped}. 
		
		\begin{figure}
			\centering
			\includegraphics[width=1\linewidth]{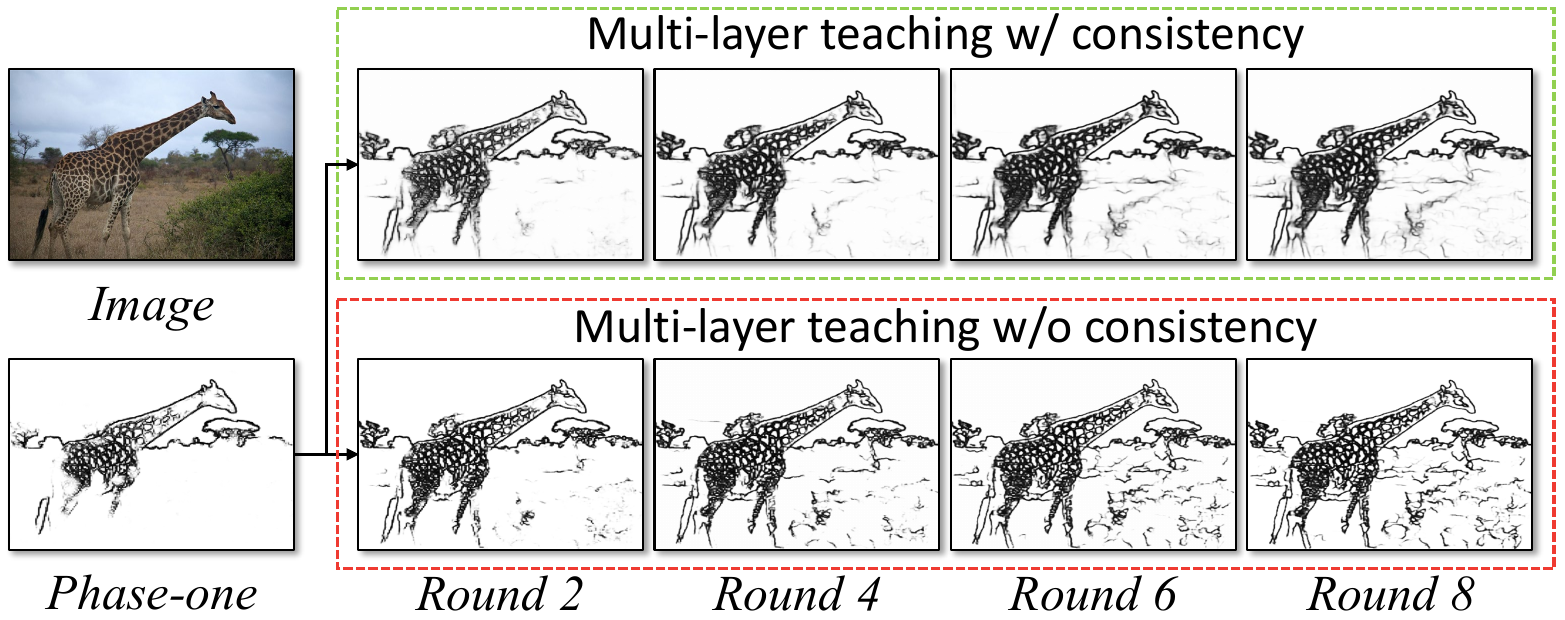}
			\caption{An example from \textit{COCO val2017} of different rounds self-trained on \textit{COCO val2017} with and without consistency. Initializing from the same phase one model, the edge maps trained with consistency learning are qualitatively much more reasonable.}
			\label{fig:consistency}
		\end{figure}
		
		As expected, it is difficult to set proper thresholds for Canny detectors to fit all images, which leads to low F-scores. However, from the noisy Canny pseudo labels, the networks successfully learn general features for edge detection. The phase-one model training from Canny labels already outperforms Canny detectors. 
		Also, we can observe that the performance drops significantly if only the last layer is supervised, which reveals the necessity of multi-layer teaching. Meanwhile, there will be much more redundant edge pixels without regularization by consistency loss. As shown in Figure~\ref{fig:consistency}, the consistency loss works as a regularizer to stop the network from labeling weak noisy edges that violate the consistencies. 
		
		Through multi-layer teaching and consistency, the networks continue to get performance boosting. Additionally, we also demonstrate that if we introduce uncertainty to further filter pseudo labels and retraining with them, the network performance will continue to improve. Note that the whole self-training process employ only \textit{unlabeled} data.
		
		
		
		Figure~\ref{fig:coco_evolve} illustrates some examples from the \textit{COCO-val2017} dataset. The initial pseudo label is sparse and noisy from Canny detector with high thresholds. 
		As iterative training proceeds, the predicted edge maps and pseudo labels are evolving together, gradually getting better. 
		
		
		\begin{table}[htbp]
			\centering
			\scalebox{1}{
				\begin{tabular}{c|ccc}
					\hline
					Canny Threshold & ODS   & OIS & AP \\
					\hline
					20-40 & .724 & .746 & .678 \\
					50-100 & .763 & .784 & .796\\
					100-200 & .759 & .782  & .777\\
					150-300 & .763 & .784 & .798\\
					200-300 & .760  & .782 & .798\\
					300-400 & .713 & .731 & 667 \\
					\hline
			\end{tabular}}
			\caption{Quantitative evaluation of self-training with unlabeled \textit{COCO-val2017} dataset and initializing with Canny of various thresholds.}
			\label{tab:canny}%
		\end{table}%
		
\begin{figure}
			\centering
			\includegraphics[width=1\linewidth]{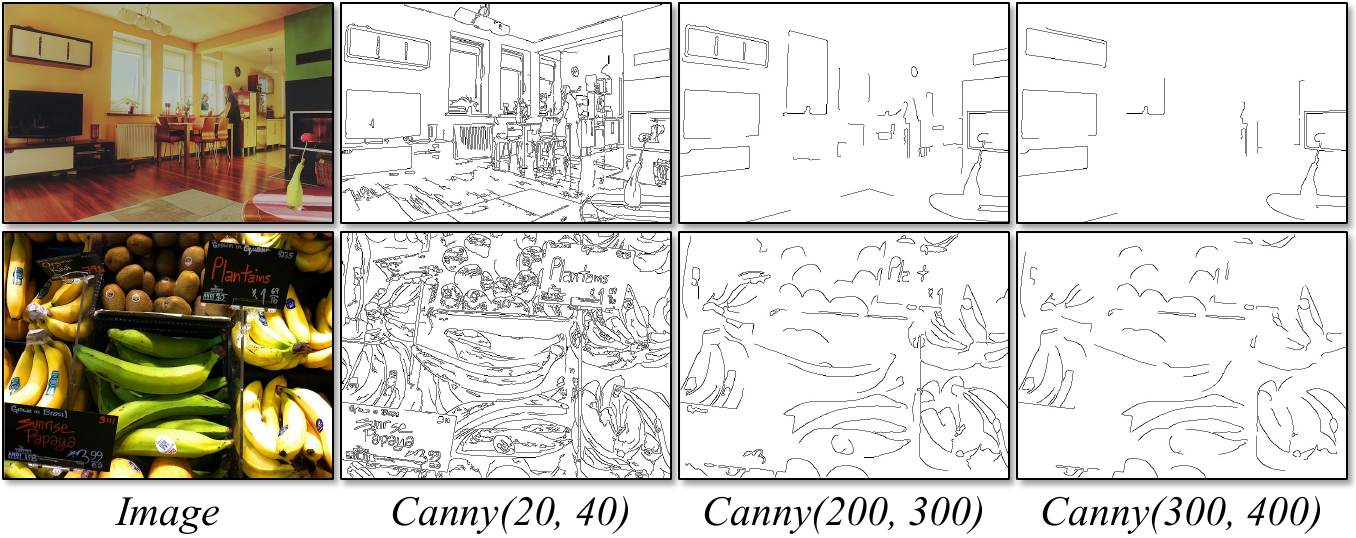}
			\caption{Examples of Canny detectors with various thresholds on \textit{COCO-val2017} dataset.}
			\label{fig:various_canny}
		\end{figure}

\begin{figure}
			\centering
			{\includegraphics[width=0.8\columnwidth]{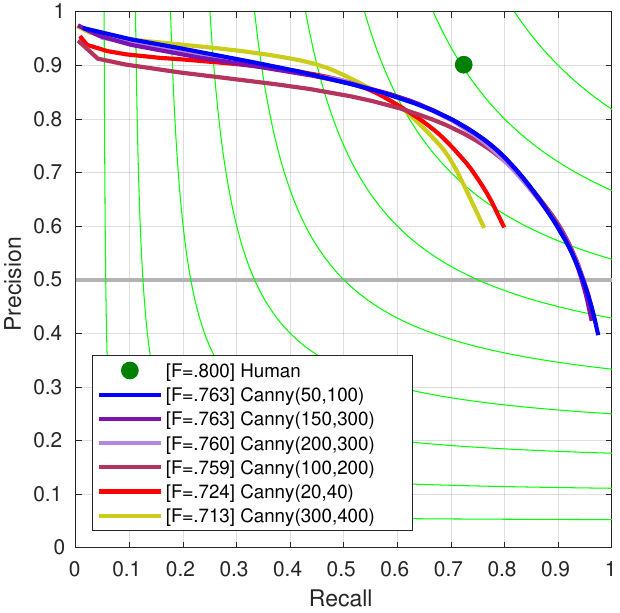}}
			\caption{The final performance self-trained from Phase-one models with different Canny thresholds. Except for some extreme cases, curves of moderate initialization have similar performances.}
			\label{fig:curve_canny}
		\end{figure}

		We also evaluate the influences of different Canny pseudo labels used in the phase-one model, quantitative results and the Precision-Recall curves are shown in Table~\ref{tab:canny} and Figure~\ref{fig:curve_canny}. When the number of edges on initial edge maps is moderate, initializing from different Canny pseudo labels does not make a big difference to the final model. However, the final performance drops if the edge pixels on initial edge maps are too sparse (300-400) or too dense (20-40). The sensitivity to the quality of initial pseudo labels is one of the limitations of our method and could be tackled by introducing extra constraints. Some visual examples are provided in Figure~\ref{fig:various_canny}. 
		Since the qualitative performance shows setting thresholds as (200,300) is slightly better, we use this setting as initialization throughout this paper.

		\begin{figure*}
			\centering
			\includegraphics[width=2\columnwidth]{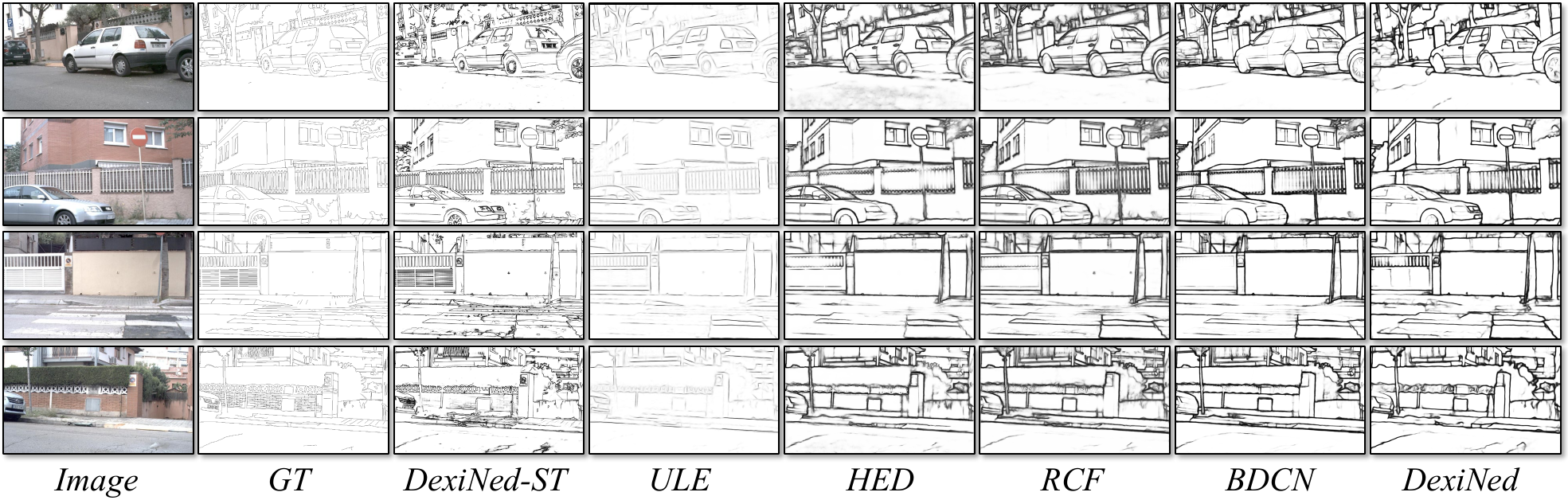}\
			\caption{Qualitative comparisons on \textit{BIPED} dataset.  ULE~\cite{li2016unsupervised} is self-trained on VSB and YTB dataset ($\sim$50K frames) and ``DexiNed-ST'' means self-trained on \textit{COCO val2017} dataset (5K images) without any human annotations using the DexiNed backbone. Other networks are trained on labeled \textit{BSDS} dataset.} 
			\label{fig:generalization}
		\end{figure*}

		\subsection{Comparisons of Generalization}
		In this section, we study the generalization ability of the proposed approach. We compare our network with several state-of-the-art networks including HED~\cite{xie2015holistically}, RCF~\cite{liu2017richer}, BDCN~\cite{he2019bi}, DexiNed~\cite{poma2020dense} and PiDiNet\cite{su2021pixel}. 
		For generalization, we train all these backbones on \textit{BSDS}, and evaluate them on \textit{BIPED} . All other methods are trained supervisedly with labeled edge maps while our network is self-trained without using any human annotations. Note that our networks do not initialized from any pretrained models  
		while others are pretrained with ImageNet~\cite{deng2009imagenet}. 
		
		Moreover, we also compare the performance with the most similar self-training edge detection framework ULE (Unsupervised Learning of Edges)~\cite{li2016unsupervised}. ULE starts with Sobel edges~\cite{kittler1983accuracy} and adopts EpicFlow~\cite{revaud2015epicflow} for optical flow estimation. The motion edges are detected on colored flow maps and then applied to train the new edge detector, which will further detect edges for the next iteration. Since optical flow estimation needs consecutive frames, ULE combines videos from two different datasets: the Video Segmentation Benchmark (VSB)~\cite{galasso2013unified} and the YouTube Object dataset (YTB)~\cite{prest2012learning}, leading to $\sim$50K frames after frame filtering.
		
\begin{figure}
			\centering
			{\includegraphics[width=0.8\columnwidth]{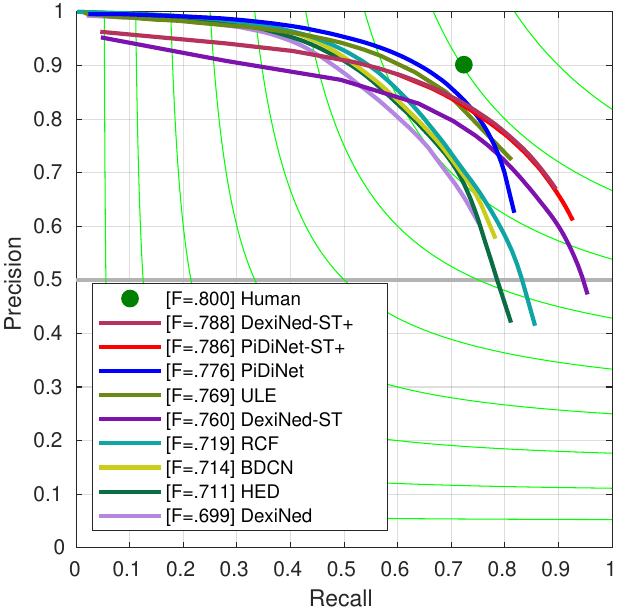}}
			\caption{Precision-recall curves of state-of-the-art methods trained on \textit{BSDS} and evaluated on \textit{BIPED}.}
			\label{fig:curve_generalization}
		\end{figure}

		The Precision-Recall curves are presented in Figure~\ref{fig:curve_generalization}.  By observing the quantitative results in Table~\ref{tab:generalization} and qualitative results in Figure~\ref{fig:generalization}, several interesting conclusions can be drawn. (a) Edge detectors with our self-training strategy outperform other methods significantly with more precise details, including those supervised on \textit{BSDS} and self-trained on larger datasets: In Table~\ref{tab:multicue}-\ref{tab:biped}, previous works are capable of performing well when training and evaluating on the same datasets. However, from Table~\ref{tab:generalization} we can see that their generalization ability is limited when evaluating on unseen datasets, showing that the generalization ability of edge detection networks remains an open problem, which is important in practical applications. ULE can detect most motion edges well, however, there also exist other types of semantic edges in natural images, which limits its performance.
		(b) Training on the pseudo labels generated by our self-training method can bring free performance boost to edge detectors: Compared with the original version, the DexiNed self-trained on unlabeled \textit{BSDS} dataset can already achieve comparable performance on unseen \textit{BIPED} dataset. The performance further boosts significantly for both the backbones of DexiNed and PiDiNet when self-trained on the \textit{COCO val} dataset, whether the network is pre-trained (on \textit{BSDS}) or not, revealing the potential of exploring more unlabeled datasets. 
		
		\begin{table}[htbp]
			\centering
			
			\scalebox{0.95}{
				\begin{tabular}{c|c|c|ccc}
					\hline
					Method &Pretraining & Training &  ODS   & OIS & AP\\
					\hline
					HED   &ImageNet & BSDS  &  .711 & .725 &.714\\
					RCF   &ImageNet & BSDS  &  .719 & .732 &.749\\
					BDCN  &ImageNet & BSDS  &  .714 & .725 &.687\\
					DexiNed &\textbackslash{} & BSDS  &  .699 & .716 &.664\\
					PiDiNet &\textbackslash{} & BSDS  &  .776 & .785 & .740\\
					\hline
					{DexiNed-ST} &\textbackslash{} &BSDS* &   .719 & .734 &.732\\
					{DexiNed-ST} &\textbackslash{} &COCO-val2017* &   .760  & .783 & \textbf{.798}\\
					{DexiNed-ST+} &\textbackslash{} & COCO-val2017* &  .788 & .803 & .754\\ 
					DexiNed-ST+ &BSDS & COCO-val2017*  &  .768 & .795 &.768\\
					
					{PiDiNet-ST+} &\textbackslash{} & COCO-val2017* & .786 & .804 &  .259\\
					PiDiNet-ST+ &BSDS & COCO-val2017*  &  \textbf{.790} & \textbf{.809} & .387\\
					
					ULE~\cite{li2016unsupervised} &\textbackslash{} &VSB*+YTB* & .769 & .782 &.742\\
					\hline
			\end{tabular}}
			\caption{Comparisons between state-of-the-art methods evaluated on \textit{BIPED}. ``-ST'' means adopting our self-training strategy based on an edge detection backbone, and ``-ST+'' means applying further uncertainty-aware retraining on \textit{COCO-val2017} (5K images) as described in Section~\ref{sec:retraining}. ULE~\cite{li2016unsupervised} is self-trained on (VSB)~\cite{galasso2013unified} and YTB~\cite{prest2012learning} datasets ($\sim$50K frames).
				\textbf{* denotes training with pseudo labels without human annotations, and same for other tables}.}
			\label{tab:generalization}%
		\end{table}%
		
		\subsection{Comparisons with State-of-the-arts}
\begin{figure*}
			\centering
			\includegraphics[width=2\columnwidth]{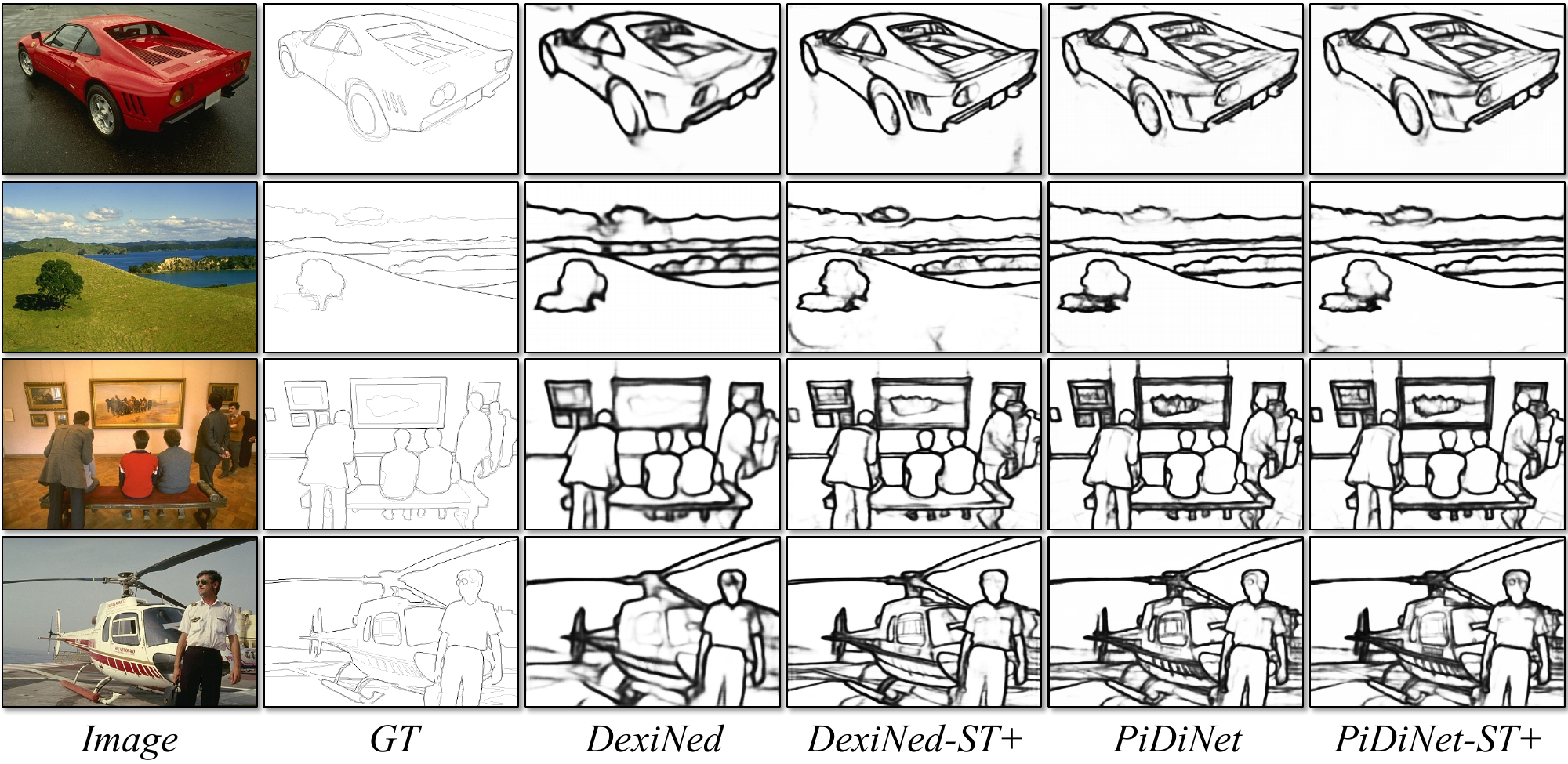}\
			\caption{Qualitative detection examples from \textit{BSDS} dataset before NMS of two backbones combined with our STEdge.``-ST+'' means applying our whole self-training pipeline with edge detection detectors, and same for other figures. With the aid of STEdge, more detailed edges are predicted precisely compared with the original ones.}
			\label{fig:bsds_backbone_and_ST}
		\end{figure*}

		{\bf{Performance on \textit{BSDS}:}} 
		We compare our model with traditional
		detectors including SCG~\cite{ren2012discriminatively}, PMI~\cite{isola2014crisp}, and OEF~\cite{hallman2015oriented}, and deep-learning-based detectors including DeepContour~\cite{shen2015deepcontour}, HED~\cite{xie2015holistically}, RCF~\cite{liu2017richer}, BDCN~\cite{he2019bi}, DexiNed~\cite{poma2020dense} and PiDiNet~\cite{su2021pixel}. The best results of all the methods are taken from their publications. 
		Some qualitative results are illustrated in Figure~\ref{fig:bsds_backbone_and_ST}. With the aid of STEdge, edge detectors can predict more detailed edges accurately.  
		Quantitative results are reported in Table~\ref{tab:bsds}. As lightweight edge detectors compared with VGG-architecture-based networks~\cite{xie2015holistically, liu2017richer, he2019bi},  DexiNed and PiDiNet with our self-training method still achieve comparable performance with much faster inference time. Also, both of them can benefits from pre-training on \textit{COCO val} dataset for free performance boost, for DexiNed, it attains $1.4\%$ improvement for ODS and $1.4\%$ for OIS, and $0.7\%$ for ODS and $0.9\%$ for OIS in terms of PiDiNet. 
		
		\begin{table}[htbp]
			\centering
			\scalebox{1}{
				\begin{tabular}{c|c|c|ccc}
					\hline
					Method & Pretraining & Training & ODS   & OIS & AP \\
					\hline
					SCG   & \textbackslash{} & BSDS & .739 & .758 & .773 \\
					PMI   & \textbackslash{} & BSDS & .741 & .769 & .799\\
					OEF   & \textbackslash{} & BSDS & .746 & .770 & .820\\
					DeepContour & \textbackslash{} & BSDS  & .757 & .776 & .800\\
					HED   & ImageNet & BSDS  & .788 & .808 & .840 \\
					RCF   & ImageNet & BSDS  & .798 & .815 &\textbackslash{} \\
					BDCN  & ImageNet & BSDS  & \textbf{.806} & \textbf{.826} & \textbf{.847}\\
					\hline
					DexiNed & \textbackslash{} & BSDS  & .760  & .779 & .690
					\\
					DexiNed-ST+ & COCO-val2017* & BSDS  & \textbf{.774} & \textbf{.793} & .770 \\
					\hline
					PiDiNet & \textbackslash{} & BSDS  & .789 & .803 & \textbackslash{}\\
					PiDiNet-ST+ & COCO-val2017* & BSDS  & \textbf{.796} & \textbf{.812} & \textbf{.796} \\
					\hline
				\end{tabular}%
			}
			\caption{Quantitative results of several recent state-of-the-art works on \textit{BSDS} dataset.}
			\label{tab:bsds}%
		\end{table}%
		
		\begin{table}[htbp]
			\centering
			\scalebox{0.9}{
				\begin{tabular}{c|c|c|ccc}
					\hline
					Method &Pretraining & Training &  ODS   & OIS &AP \\
					\hline
					Human &\textbackslash{} & \textbackslash{} &  .750  & \textbackslash{} & \textbackslash{}\\
					Multicue &\textbackslash{} & Multicue &  .830  & \textbackslash{} & \textbackslash{}\\
					HED   &ImageNet & Multicue &  .851 & .864 & .890\\
					RCF   &ImageNet & Multicue &  .857 & .862 & \textbackslash{}\\
					BDCN  &ImageNet & Multicue &  .891 & .898 &.935\\
					\hline
					DexiNed &\textbackslash{} & Multicue &  .872 & .881 &.871\\
					DexiNed-ST+ &COCO-val2017* & Multicue 50\% &  .889 & .896 &.943\\
					DexiNed-ST+ & COCO-val2017* & Multicue &  \textbf{.895} & \textbf{.900} &\textbf{.948}\\
					\hline
					PiDiNet &\textbackslash{} & Multicue &  .855 & .860 &\textbackslash{}\\
					PiDiNet-ST+ &COCO-val2017* & Multicue 50\% &  .878 & .885 &.903\\
					PiDiNet-ST+ & COCO-val2017* & Multicue &  \textbf{.881} & \textbf{.887} &.901\\
					\hline
			\end{tabular}}
			\caption{Comparison with state-of-the-art works on \textit{Multicue} dataset. 50\% means finetuning using half of the training set.}
			\label{tab:multicue}%
		\end{table}%
		
\begin{figure*}
			\centering
			\includegraphics[width=2\columnwidth]{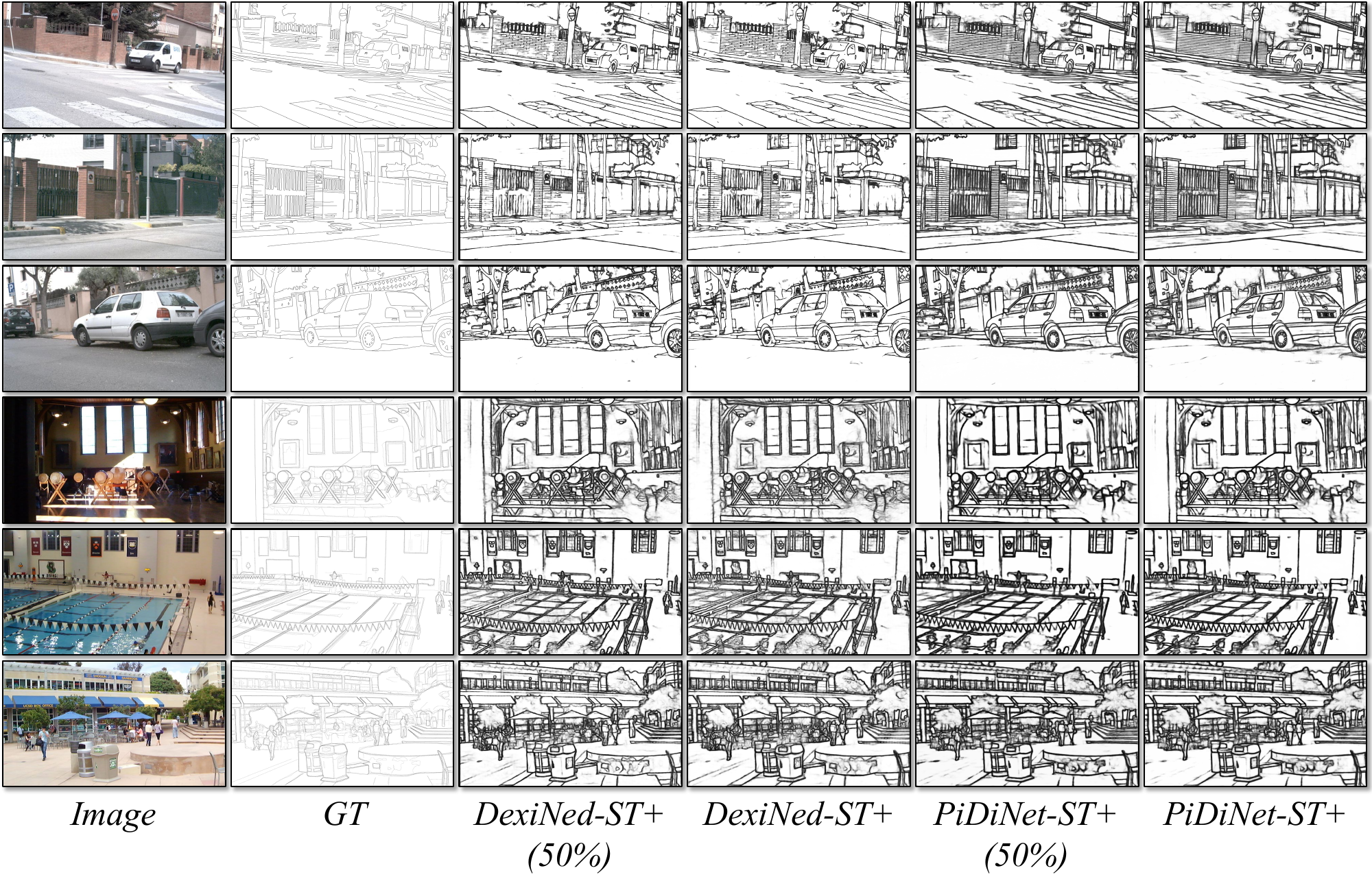}\
			\caption{Qualitative detection results before NMS of two backbones combined with our STEdge. The first three rows are examples from \textit{BIPED} dataset and the last three rows come from \textit{Multicue} dataset. ``(50\%)'' means training using half of the training set. With the aid of our STEdge, edge detectors finetuned only on half of the training set already achieve state-of-the-art performances.}
			\label{fig:multicue_biped_backbone_and_ST}
		\end{figure*}

		{\bf{Performance on \textit{Multicue}:}} 
		For a fair comparison, we follow the experiments of previous works~\cite{liu2017richer,he2019bi}, finetuning our self-trained model on the split 80\% training set and test on the remaining 20\%. 
		We average the scores of three independent trials as the final results. 
		Some Qualitative examples are presented in Figure~\ref{fig:multicue_biped_backbone_and_ST}. After self-training on \textit{COCO val2017}, even finetuned on $50\%$ of the training set, the edge detectors can achieve almost the same performances compared with those finetuned on the whole training set, showing the potential of our self-training method for few-shot edge detection.
		The comparisons to recent state-of-the-art methods are reported in Table~\ref{tab:multicue}, where our STEdge with DexiNed backbone achieves the best performance. Moreover, we can observe that, for both DexiNed and PiDiNet with STEdge, even finetuned on $50\%$ of the training set, their performance already outperform most of previous methods and their original performance without self-training on \textit{COCO val}. Especially, if trained with the same $100\%$ training set, with the aid of our self-training strategy, DexiNed attains $2.3\%$ improvement for ODS and $1.9\%$ for OIS, PiDiNet attains $2.6\%$ improvement for ODS and $2.7\%$ for OIS.

		\noindent{\bf{Performance on \textit{BIPED}:}} \textit{BIPED} is a recent dataset with well-annotated edge maps.
		We also finetune the self-trained models on $50\%$ and $100\%$ of the training split, which correspond to 100 and 200 images respectively. As shown in Table~\ref{tab:biped}, DexiNed and PiDiNet with STEdge method also already achieve the best performance. With more labeled images($100\%$) for training, the performance gets substantially higher. By self-training on \textit{COCO val} dataset freely, DexiNed attains $0.7\%$ improvement for ODS and $1.6\%$ for OIS, PiDiNet attains $1\%$ improvement for ODS and $0.7\%$ for OIS.
		
		\begin{table}[htbp]
			\centering
			\scalebox{0.95}{
				\begin{tabular}{c|c|c|ccc}
					\hline
					Method &\multicolumn{1}{c|}{Pretraining} & Training &  ODS   & OIS &AP\\
					\hline
					HED   &\multicolumn{1}{c|}{ImageNet} & BIPED &  .829 & .847 &.869\\
					RCF   &\multicolumn{1}{c|}{ImageNet} & BIPED &  .843 & .859 &.882\\
					BDCN  &\multicolumn{1}{c|}{ImageNet} & BIPED &  .839 & .854 &.887\\
					\hline
					DexiNed &\multicolumn{1}{c|}{\textbackslash{}} & BIPED &  .857 & .861 &.805\\
					DexiNed-ST+ &COCO-val2017* & BIPED 50\% & .859 & .873 &.737
					\\
					DexiNed-ST+ &COCO-val2017* & BIPED &  \textbf{.867} & \textbf{.878} & .732
					\\
					\hline
					PiDiNet &\multicolumn{1}{c|}{\textbackslash{}} & BIPED &  .868 & .876 & .912\\
					PiDiNet-ST+ &COCO-val2017* & BIPED 50\% & .876 & .881 &.913
					\\
					PiDiNet-ST+ &COCO-val2017* & BIPED &  \textbf{.878} & \textbf{.883} & \textbf{.923}\\
					\hline
			\end{tabular}}
			\caption{Comparison with state-of-the-art works on \textit{BIPED} dataset. 50\% means finetuning using half of the training set,and same for other tables.}
			\label{tab:biped}%
		\end{table}%

		The experiments demonstrate the superiority of the proposed self-training scheme and multi-layer consistency regularization. By exploring unlabeled datasets, the generalization ability and performance are both improved significantly.

		\section{Conclusion and Limitations}
		We propose a simple but effective self-training framework for edge detection named STEdge, to leverage unlabeled image datasets. To the best of our knowledge, it is among the first self-training pipeline proposed for edge detection. 
		The framework consists of multi-layer teaching by noisy pseudo labels and consistency regularization to suppress complicated textures. During iterative training, network predictions and the noisy pseudo labels are evolving simultaneously. 
		Experimental results show the superiority of the proposed STEdge on generalization ability and edge detection performance on several benchmark datasets. 
		In the future, we plan to explore the full pipeline of utilizing unlabeled web images for self-training edge detection and further improve the cross-dataset generality. 
		
		\textit{Limitations}: The performance drops when the initial edge maps are too sparse or too dense, the initialization strategy with designed constraints could be one of the future direction. Although applying STEdge to edge detectors can achieve better performances, the utilization of unlabeled data is still unsatisfactory. Some pseudo labels still contain noisy pixels surrounding image areas of complicated textures, degenerating the self-training performance. 
		A well-designed pseudo label sampling strategy during re-training is also an interesting direction to explore.
		
		
		
		
		
		


		
		%
		
		\newpage
		\bibliographystyle{IEEEtran}
		\bibliography{references}

		
		%
		%
		%
		%

		\vfill
		
	\end{document}